\pdfoutput=1

\documentclass[11pt]{article}

\usepackage{emnlp2021}

\usepackage{times}
\usepackage{latexsym}

\usepackage[T1]{fontenc}

\usepackage[utf8]{inputenc}
\usepackage{adjustbox}
\usepackage{microtype}
\usepackage{float}
\usepackage{graphicx}
\usepackage{multirow}
\usepackage{adjustbox}
\DeclareUnicodeCharacter{0002}{ }
\title{Improving Machine Translation with Phrase Pair Injection and Corpus Filtering}

\author{Akshay Batheja, Pushpak Bhattacharyya\\[1ex]
CFILT, Indian Institute of Technology Bombay \\
\texttt{\{akshaybatheja, pb\}@cse.iitb.ac.in}
}
\begin{document}
\maketitle
\begin{abstract}
    In this paper, we show that the combination of Phrase Pair Injection and Corpus Filtering boosts the performance of Neural Machine Translation (NMT) systems. We extract parallel phrases and sentences from the pseudo-parallel corpus and augment it with the parallel corpus to train the NMT models. With the proposed approach, we observe an improvement in the Machine Translation (MT) system for 3 low-resource language pairs, Hindi-Marathi, English-Marathi, and English-Pashto, and 6 translation directions by up to 2.7 BLEU points, on the FLORES test data. These BLEU score improvements are over the models trained using the whole pseudo-parallel corpus augmented with the parallel corpus. 



    
\end{abstract}



\section{Introduction}
Deep Neural Architectures are \textit{data hungry}. It is believed that the robustness of an NMT model depends on the size of the training corpus. However, not all language pairs have a substantial amount of parallel data. The primary data resource for low-resource languages is the web. However, the web-crawled data contains a lot of noise that degrades the performance of NMT systems. Hence, the quality of the training data is as important as its quantity. We aim to improve the quality of Machine Translation for Hindi-Marathi, English-Marathi, and English-Pashto language pairs by using the LaBSE-based \cite{labse} corpus filtering along with the Phrase Pair Injection. We use Phrase Pair Injection to increase the size and LaBSE-based corpus filtering to improve the quality of the parallel data extracted from the pseudo-parallel corpus. We observe that using these two techniques together makes the optimum use of the pseudo-parallel corpus.\\
The contributions of this work are as follows:
\begin{itemize}
    \item We extract good quality parallel sentences and phrases from the pseudo-parallel corpus of huge size. To the best of our knowledge, this is the first method that combines Phrase Pair Injection with neural-based Corpus Filtering and extracts both good quality parallel sentences and phrases from the pseudo-parallel corpus. 
    \item We show that the extracted parallel sentences and phrases significantly improve the performance of the MT systems.
   
\end{itemize}

\section{Related Work}
Neural Networks have become very popular with increased computational power in recent times. The Transformer model introduced by \citet{vaswani} gave significant improvements in the quality of translation as compared to the previous approaches \cite{bahdanau,Sutskever}. Transformers allow parallelization, which enables the model to train faster and get better performances.\\
\indent There are several prior studies to improve the quality and size of the parallel corpus. A set of heuristic rules were proposed by \citet{alibaba2} to remove low-quality sentence pairs from the noisy parallel corpus. Another popular approach is to compute cosine similarities between the source and target sentence embeddings.  \citet{labse} proposed the LaBSE model, which is a multilingual sentence embedding model trained on 109 languages, including some Indic languages. \citet{pindomain} proposed a data selection pipeline that selects the In-Domain data from the Generic Domain based on its similarity with the other In-Domain data. This helps the MT model to perform well in domain-specific cases. \\
\indent \citet{sen_hasanuzzaman_ekbal_bhattacharyya_way_2021} augmented the raw phrase table with the parallel corpus. The raw phrase table contains very noisy and repetitive phrase pairs. We observe that augmenting the whole phrase table with parallel corpus does not show much improvement in the performance of the MT system.
In contrast, we first extract the longest unique phrase pairs from the phrase table and then further filter them using LaBSE filtering to extract good quality phrase pairs. Then we augment these good quality phrase pairs with the LaBSE filtered parallel sentences. This helps improve the performance of MT systems significantly.

\begin{figure}[t]
\begin{center}
\includegraphics[scale=.4]{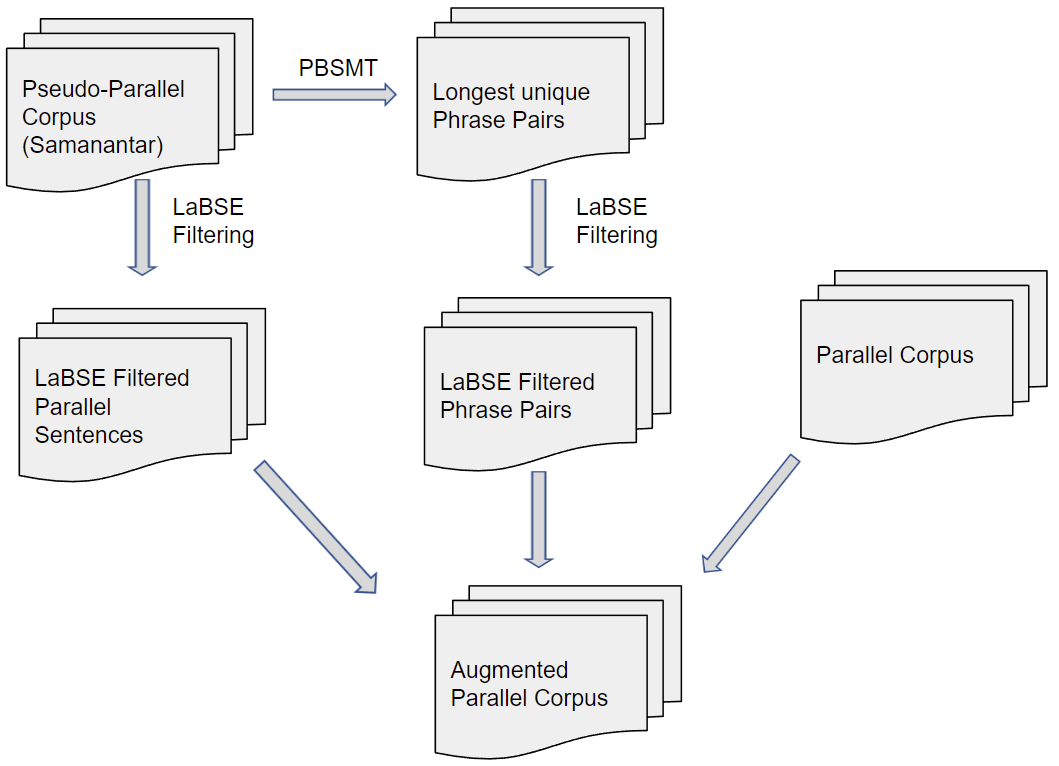}\caption{Phrase Pair Injection with LaBSE Filtering Pipeline}
\label{fig:1}
\end{center}
\end{figure}

\section{Approaches}
We first discuss a method to extract good-quality parallel sentences from the pseudo-parallel corpus. Then we discuss a method to extract good quality phrase pairs from the pseudo-parallel corpus.

\subsection{LaBSE based Filtering}
In this approach, we aim to extract good-quality parallel sentences from the pseudo-parallel corpus. Language Agnostic BERT Sentence Embedding model \cite{labse} is a multilingual embedding model that supports 109 languages, including some Indic languages. We generate the sentence embeddings for the source and target sides of the pseudo-parallel corpora using the LaBSE \footnote{\url{https://huggingface.co/sentence-transformers/LaBSE}} model.
Then, we compute the cosine similarity between the source and target sentence embeddings. After that, we extract good quality parallel sentences based on a threshold value of the similarity scores. We calculate the average similarity score on a small dataset from the PM-India corpus (PMI) \cite{pmi}. The PMI corpus consists of high-quality sentence pairs, so it helps us decide upon the threshold value.

\subsection{Phrase Pair Injection (PPI) with LaBSE Filtering}

In this approach, we aim to extract good-quality parallel phrases from a noisy pseudo-parallel corpus. We first train a PBSMT model on the noisy pseudo-parallel corpus using the \textit{Moses} \footnote{\url{http://www2.statmt.org/moses/?n=Development.GetStarted}} decoder, followed by tuning using MERT. Then,  we extract phrase pairs from the generated Phrase Table based on their weighted average translational and lexical probabilities mentioned in the phrase table. As each sentence pair leads to multiple and repetitive phrase pairs in the Phrase Table, we only keep the longest unique phrase pairs from the extracted phrase pairs. Then, we perform the LaBSE-based filtering on these longest unique phrase pairs to remove the poor quality phrase pairs.\\
\indent We augment these good-quality phrase pairs to the LaBSE filtered parallel sentences. 
This approach allows us to use the pseudo-parallel corpora at its full potential because we are extracting good quality parallel sentences as well as phrase pairs. 


\begin{table}[htp]
\centering
\begin{adjustbox}{width=\columnwidth,center}
\begin{tabular}{l|l|r}
\hline
\textbf{Corpus Name} & \textbf{Language} &\textbf{Sentence}\\
&\textbf{Pairs} & \textbf{Pairs}\\
\hline
\multirow{3}{*}{\textbf{Parallel Corpus}} &Hindi-Marathi& 604K \\
&English-Marathi& 248K\\ 

&English-Pashto& 123K \\ \hline

\multirow{3}{*}{\textbf{Pseudo-Parallel Corpus}} &Hindi-Marathi&1.98M\\
 &English-Marathi&3.28M\\
 &English-Pashto&1.02M\\\hline
\end{tabular}
\end{adjustbox}
\caption{Dataset Statistics of  Parallel and Pseudo-Parallel Corpus}
\label{lab:corpus}
\end{table}



\begin{table*}[htp]
    
    \small
    \centering
    \begin{adjustbox}{width=\textwidth,center}
    \begin{tabular}{lrrrrrrrrr}
     \hline
     \textbf{Technique}& \textbf{\# Sentence} & 
     \multicolumn{4}{c}{\textbf{Hindi$\rightarrow$Marathi}}& \multicolumn{4}{c}{\textbf{Marathi$\rightarrow$Hindi}}\\
     &  \textbf{Pairs}& \multicolumn{2}{c}{\textbf{WAT 2021}}  & \multicolumn{2}{c}{\textbf{FLORES 101}}& \multicolumn{2}{c}{\textbf{WAT 2021}}  & \multicolumn{2}{c}{\textbf{FLORES 101}}\\
     & & \textbf{BLEU} & \textbf{chrF} & \textbf{BLEU} & \textbf{chrF}&\textbf{BLEU} & \textbf{chrF}&\textbf{BLEU} & \textbf{chrF}\\  
     \hline
      \textbf{Baseline+LaBSE +PPI- LaBSE}&1.75M&\textbf{17.6}& \textbf{51.0} & 10.0& 42.5 & \textbf{25.4}& 51.5 &16.2&42.5 \\
         \textbf{Baseline+LaBSE +PPI}& 2.55M&17.4&51.0 &\textbf{10.2}& \textbf{43.0} &25.3& \textbf{51.6} & \textbf{16.5}& \textbf{42.5} \\
     \textbf{Baseline+PPI-LaBSE}& 1.4M&15.5 &49.8 & 9.5&42.0 &23.4& 50.2  &15.4& 41.6\\
     \textbf{Baseline+LaBSE}& 960K &17.5 & 50.3 &9.4&41.9 &25.1& 51.2 &15.9& 41.6  \\
     \textbf{Baseline+PPI}&2.2M&16.1& 49.9 &9.2&41.4 &23.1&49.9 & 15.1&40.8 \\
     \textbf{No Filtering}& 2.56M&16.8& 49.9 &9.1& 39.5 & 24.0& 49.9 &14.8& 39.8  \\
     \textbf{Baseline}&604K&15.2&48.0 &7.6& 38.8 &22.1& 48.1  &14.4& 39.7 \\
     
    \hline
    \end{tabular}
    \end{adjustbox}
    \caption{\textbf{BLEU} and \textbf{chrF} scores of Hindi-Marathi NMT models.}
     
    \label{tab:resulta}
\end{table*}

     
     

\begin{table*}[t]
    \small
    \centering
    \begin{tabular}{ p{4.5cm}rrrrr}
     \hline
     \textbf{Technique}& \textbf{\# Sentence} & \multicolumn{2}{c}{\textbf{English$\rightarrow$Marathi}} & \multicolumn{2}{c}{\textbf{Marathi$\rightarrow$English}}\\ &\textbf{Pairs} &\multicolumn{2}{c}{\textbf{FLORES 101}}&\multicolumn{2}{c}{\textbf{FLORES 101}}\\
     & & \textbf{BLEU} & \textbf{chrF} & \textbf{BLEU} & \textbf{chrF}\\
     \hline
     \textbf{Baseline + LaBSE + PPI-LaBSE}&  3.24M & 9.8 & 40.3 &\textbf{17.0}& \textbf{44.6} \\
     \textbf{Baseline + LaBSE + PPI}&  4.09M&\textbf{9.9}& \textbf{40.5}&16.2&43.4 \\
     \textbf{Baseline + PPI-LaBSE}&  640K &6.2& 33.8 &12.7& 40.3 \\
     \textbf{Baseline + LaBSE}&2.85M &8.8& 39.8 &16.7& 44.0\\
     \textbf{Baseline + PPI}&  1.49M &6.6& 35.0 &12.7&40.3 \\
     \textbf{No Filtering}&  3.53M &8.8& 37.8&15.9& 43.3\\
     \textbf{Baseline}& 248K &5.1& 32.4 &10.2&38.0\\
     \hline
    \end{tabular}
    
    \caption{\textbf{BLEU} and \textbf{chrF} scores of English-Marathi NMT models}
    \label{tab:resultb}
\end{table*}

\begin{table*}[t]
    \small
    \centering
    \begin{tabular}{p{5.5cm}rrrrr}
     \hline
     \textbf{Technique}& \textbf{\# Sentence} & \multicolumn{2}{c}{\textbf{English$\rightarrow$Pashto}} & \multicolumn{2}{c}{\textbf{Pashto$\rightarrow$English}}\\ & \textbf{Pairs}&\multicolumn{2}{c}{\textbf{FLORES 101}}&\multicolumn{2}{c}{\textbf{FLORES 101}}\\
     & & \textbf{BLEU} & \textbf{chrF} & \textbf{BLEU} & \textbf{chrF}\\
     \hline
     \textbf{Baseline + LaBSE + PPI-LaBSE}&  342K &8.6& 31 & 10.0& 36.2\\
     \textbf{Baseline + LaBSE + PPI}&  790K &\textbf{8.7}& \textbf{31} &\textbf{10.4}& \textbf{37.1}\\
     \textbf{Baseline + PPI-LaBSE}&  176K & 0.9 & 13.1 &1.1& 17.3 \\
     \textbf{Baseline + LaBSE}& 290K &8.0& 30.5& 9.8 &36.4\\
     \textbf{Baseline + PPI}&  624K &0.7&12.3 &0.8& 13.9\\
     \textbf{No Filtering}&  1.14M&6.0& 23.2&9.4& 34.4\\
     \textbf{Baseline}& 124K &0.2& 8.7 &0.4& 15.9\\
     \hline
    \end{tabular} 
    
    \caption{\textbf{BLEU} and \textbf{chrF} scores of English-Pashto NMT models}
    \label{tab:resultb}
\end{table*}

\section{Experimental Setup}
In this section, we discuss the setup of various experiments that we performed. We use Byte Pair Encoding \cite{sennrich-etal-2016-neural} as a segmentation technique to split words into subwords. We use 16000 merge operations for all experiments. We use the \textit{OpenNMT-py}\footnote{\url{https://github.com/OpenNMT/OpenNMT-py}} \cite{opennmt} library to train the Transformer based NMT models. We also train the PBSMT systems using Moses \cite{Koehn2007MosesOS}. Then, we perform tuning using Minimum Error Rate Training (MERT) to find the optimal weights which maximize the translation performance. We use a tune set of the first 2000 parallel sentences from the pseudo-parallel corpus for the MERT tuning. Then, we extract the phrase pairs from the generated Phrase Table, based on their weighted average translational and lexical probabilities mentioned in the phrase table. Further training details can be found in the appendix \ref{sec:appendix1}.   
\subsection{Dataset Preparation}
We train the Hindi-Marathi, English-Marathi, and English-Pashto NMT models using their respective Parallel and Pseudo-Parallel Corpus. The Parallel corpus for English-Marathi and Hindi-Marathi language pairs consists of the Indian Languages Corpora Initiative (ILCI) phase 1  corpus \cite{ilci}, BIBLE corpus \cite{bible}, Press Information Bureau corpus (PIB), and PM-India corpus (PMI) \cite{pmi}. The Hindi-Marathi Parallel corpus also consists of Tatoeba challenge dataset \cite{tatoeba}. In the Hindi-Marathi Parallel corpus, except for ILCI, the parallel data was synthetically generated by translating the English sentences of the English-Marathi corpus to Hindi using Google Translation API. The Pseudo-Parallel corpus consists of Samanantar Corpus \cite{samanantar}. For English-Pashto language pair, we use the parallel and pseudo-parallel corpus provided by the WMT20 shared task on Parallel Corpus Filtering and Alignment \cite{koehn-etal-2020-findings}. The detailed corpus statistics are mentioned in Table \ref{lab:corpus}.\\
\indent For evaluation, we use the test set introduced in WAT 2021 MultiIndicMT: An Indic Language Multilingual Task and FLORES 101.  The test set from WAT 2021 contains 2,390 sentences and is a part of the PMI corpus. The FLORES 101 test set contains 1012 sentences.
\subsection{Baseline}\label{baseline}
We train the baseline models (Hindi-Marathi, English-Marathi and English-Pashto) directly on their respective parallel corpus.
\subsection{No Filtering}\label{nofilter}
In No Filtering model, we train the NMT models on the whole pseudo-parallel corpus augmented with parallel corpus.

\subsection{Baseline + PPI}\label{baseppi}
In this model, we first train a PBSMT model on the pseudo-parallel Corpus, followed by tuning using MERT. Then, we extract longest unique phrase pairs from the Phrase Table using a threshold Probability value of 0.95 for Hindi-Marathi and 0.8 for English-Marathi. Finally, we augment the parallel corpus with the extracted phrases pairs to train the NMT models.

\subsection{Baseline + LaBSE}\label{baselabse}
In this model,  we use the LaBSE filtering technique with a threshold of \textbf{0.9}, \textbf{0.8} and \textbf{0.8} to extract 350K, 2.6M, and 166K good-quality parallel sentences from the Hindi-Marathi, English-Marathi and English-Pashto pseudo-parallel corpus, respectively. Then, we augment the parallel corpus with the LaBSE filtered parallel sentences and train the respective NMT models.

\subsection{Baseline + LaBSE + PPI}
In this model, we combine the two techniques mentioned in the section \ref{baseppi} and \ref{baselabse}. We first extract longest unique phrase pairs from pseudo-parallel corpus. Then, we extract good quality parallel sentences from pseudo-parallel corpus using LaBSE filtering. Finally, we augment the parallel corpus with the extracted longest unique phrase pairs and LaBSE filtered parallel sentences to train the NMT models. 
\subsection{Baseline + PPI-LaBSE}\label{baseppilabse}
In this model, we first extract the longest unique phrase pairs using the technique mentioned in section \ref{baseppi}. Then we apply LaBSE filtering on these extracted phrase pairs using a threshold value of 0.9. Finally, we augment the parallel corpus with the LaBSE filtered phrase pairs to train the NMT models. 
\subsection{Baseline + LaBSE + PPI-LaBSE}
In this model, We combine the techniques mentioned in the section \ref{baselabse} and \ref{baseppilabse}. We first extract LaBSE filtered parallel sentences from pseudo-parallel corpus. Then, we extract LaBSE filtered longest unique phrase pairs from the pseudo-parallel corpus. Finally, we augment the parallel corpus with the LaBSE filtered parallel sentences and LaBSE filtered phrase pairs.

\section{Results and Analysis}
We evaluate our NMT models using BLEU \cite{bleu} and \textbf{chrF} score metric. We use sacrebleu \cite{sacrebleu} python library to calculate the BLEU and chrF scores. The results of all the experiments are summarized in Table \ref{tab:resulta}, \ref{tab:resultb}.\\
\indent We can observe from the results that the models trained using the extracted phrase pairs and parallel sentences outperform the models trained using the whole pseudo-parallel corpus when augmented with the parallel corpus. The results of Hindi-Marathi, English-Marathi and English-Pashto MT models show that \textbf{Baseline $+$ LaBSE $+$ PPI-LaBSE} and \textbf{Baseline $+$ LaBSE $+$ PPI}, perform best amongst others on \textbf{WAT 2021} and \textbf{FLORES 101} test data, respectively.\\
\indent In Hindi-Marathi MT, the best models improve the performance by \textbf{0.8} and \textbf{1.1} BLEU score points in \textit{Hindi$\rightarrow$Marathi} and  \textbf{1.4} and \textbf{1.7} BLEU score points in \textit{Marathi$\rightarrow$Hindi} over the \textbf{No Filtering} model on FLORES 101 and WAT2021 test data. \\
\indent In English-Marathi MT, the best models improve the performance by \textbf{1.0} and \textbf{1.1} BLEU score points in \textit{English$\rightarrow$Marathi} and \textit{Marathi$\rightarrow$English} over the \textbf{No Filtering } model, on the FLORES 101 test data. \\
\indent In English-Pashto MT, the best models improve the performance by \textbf{2.7} and \textbf{1.0} BLEU score points over the \textbf{No Filtering} model, on the FLORES 101 test data.\\
\indent We observe that even though the corpus size of the best models is smaller than the pseudo-parallel corpus, it still performs better as it has a higher proportion of good quality parallel sentences and phrase pairs.

\section{Conclusion and Future Work}
In this work, we show that extracting parallel sentences and phrase pairs from a pseudo-parallel corpus helps the NMT models improve its performance significantly when augmented with a parallel corpus. We show that LaBSE filtering assists the Phrase Pair Injection to extract the parallel data, which has good quality and quantity.\\
\indent In the future, we plan to use the proposed corpus filtering techniques for other language pairs. This will provide us with a general overview of how these filtering techniques perform on multiple languages. We also anticipate improvements in the result by trying different threshold values in the LaBSE filtering and Phrase Pair Injection.
\section{Acknowledgements}
We would like to thank the anonymous reviewers for their insightful feedback. We also express our gratitude towards Shivam Mhaskar, Sourabh Deoghare, Jyotsana Khatri, and other members of the Machine Translation group at CFILT, IIT Bombay, for their interesting and insightful comments.
\section{Limitations}
The size of extracted phrase pairs from the pseudo-parallel corpus is huge, and computing sentence embeddings using LaBSE for such a huge corpus is computationally expensive. We use various threshold values in LaBSE Filtering and Phrase Pair Injection. We need to try multiple combinations of these threshold values to obtain good results. 

\bibliographystyle{acl_natbib}
\bibliography{custom}

\appendix

\section{Models}
\label{sec:appendix1}
We use a Transformer based architecture to train Hindi-Marathi and English-Marathi NMT models for all our experiments. The encoder of the Transformer consists of 6 encoder layers and 8 encoder attention heads. The encoder uses embeddings of dimension 512. The decoder of the Transformer also consists of 6 decoder layers and 8 decoder attention heads.

\section{Training Details}
\label{sec:appendix2}
We use the OpenNMT-py library to train the Transformer based NMT models. The hyperparameter values are selected using manual tuning. The optimizer used was adam with betas (0.9, 0.98). The initial learning rate used was 5e-4 with the inverse square root learning rate scheduler. We use 8000 warmup updates. The dropout probability value used was 0.1 and the criterion used was label smoothed cross entropy with label smoothing of 0.1. We use batch size of 4096 tokens. All the models were trained for 2,00,000 training steps.\\
\indent We use Nvidia A100 GPUs with 40 GB memory to train our NMT models. The average training time of the models is 8 hours. The model parameters of each experiment are between 65M and 66M. The model parameters of the best models are given in the following section. 
\subsection{Baseline + LaBSE + PPI-LaBSE}
In this experiment, the Hindi-Marathi and English-Marathi models consist of 66,263,552 (66.2M) and 65,812,992 (65.8M) parameters, respectively. We perform augmentation by providing the parallel corpus and extracted good quality parallel data to the OpenNMT-py tool as two corpora with the same weight (by default weight=1).
\subsection{Baseline + LaBSE + PPI}
In this experiment, the Hindi-Marathi and English-Marathi models consist of 66,348,544 (66.3M) and 66,554,368 (66.5M) parameters, respectively.
\end{document}